\documentclass{article}
\usepackage{ijcai23}

\usepackage{times}
\usepackage{soul}
\usepackage{url}
\usepackage[hidelinks]{hyperref}
\usepackage[utf8]{inputenc}
\usepackage[small]{caption}
\usepackage{graphicx}
\usepackage{amsmath}
\usepackage{amsthm}
\usepackage{booktabs}
\usepackage[linesnumbered,ruled]{algorithm2e}
\usepackage[switch]{lineno}

\usepackage{hyperref}
\usepackage{url}
\usepackage{bm}
\usepackage{amsfonts}
\usepackage{graphics}
\usepackage{subcaption}
\usepackage{multirow}
\usepackage{amssymb}
\usepackage{wrapfig}
\usepackage{float}
\usepackage{makecell}
\usepackage{diagbox}
\usepackage{color, colortbl}
\definecolor{Gray}{gray}{0.9}
\usepackage{adjustbox}
\usepackage{pifont}
\usepackage{xcolor}
\usepackage{bbding}       % \Checkmark,\XSolid,... (需要和pifont宏包共同使用)

\definecolor{brilliantrose}{rgb}{1.0, 0.33, 0.64}
\newcommand{\xmark}{\ding{55}}%
\definecolor{lightgrey}{HTML}{dcdbdb}
\newtheorem{theorem}{Theorem}

\title{ProMix: Combating Label Noise via Maximizing Clean Sample Utility}

\author{
Ruixuan Xiao$^1$\and
Yiwen Dong$^1$\and
Haobo Wang$^1$\footnote{Corresponding Author.}\and
Lei Feng$^2$\and \\ 
Runze Wu$^3$\and
Gang Chen$^1$\And
Junbo Zhao$^1$
\affiliations
$^1$Zhejiang University, Hangzhou, China \\ $^2$Nanyang Technological University, Singapore \\$^3$NetEase Fuxi AI Lab, Hangzhou, China
\emails
\{xiaoruixuan, dyw424, wanghaobo, cg, j.zhao\}@zju.edu.cn, \\lfengqaq@gmail.com, wurunze1@corp.netease.com
}

\begin{document}

\maketitle

\begin{abstract}
    Learning with Noisy Labels (LNL) has become an appealing topic, as imperfectly annotated data are relatively cheaper to obtain. Recent state-of-the-art approaches employ specific selection mechanisms to separate clean and noisy samples and then apply Semi-Supervised Learning (SSL) techniques for improved performance. However, the selection step mostly provides a medium-sized and decent-enough clean subset, which overlooks a rich set of clean samples. To fulfill this, we propose a novel LNL framework ProMix that attempts to maximize the utility of clean samples for boosted performance. Key to our method, we propose a matched high confidence selection technique that selects those examples with high confidence scores and matched predictions with given labels to dynamically expand a base clean sample set. To overcome the potential side effect of excessive clean set selection procedure, we further devise a novel SSL framework that is able to train balanced and unbiased classifiers on the separated clean and noisy samples. Extensive experiments demonstrate that ProMix significantly advances the current state-of-the-art results on multiple benchmarks with different types and levels of noise. It achieves an average improvement of 2.48\% on the CIFAR-N dataset. The code is available at \href{https://github.com/Justherozen/ProMix}{\color{brilliantrose}{https://github.com/Justherozen/ProMix}}.
\end{abstract}

\section{Introduction}
The great success of deep neural networks (DNNs) attributes to massive and accurately-labeled training data, which is notoriously expensive and time-consuming to obtain. As an alternative, online queries \cite{DBLP:journals/jacm/BlumKW03} and crowdsourcing \cite{DBLP:journals/ml/YanRFRD14} have been widely used to enhance the labeling efficiency. However, due to the lack of expertise, such annotation processes inevitably introduce wrong labels to models and result in performance degradation. To this end, learning with noisy labels (LNL) \cite{DBLP:conf/iclr/ZhangBHRV17}, which aims to mitigate the side-effect of label noise, has attracted huge attention from the community.

A plethora of works has been developed for the LNL problem. Some early approaches 
design noise-robust loss \cite{DBLP:conf/iccv/0001MCLY019,https://doi.org/10.48550/arxiv.2212.04055}. While some other classical methods utilize the latent noise transition matrix to calibrate the loss function \cite{DBLP:conf/cvpr/PatriniRMNQ17,DBLP:conf/nips/XiaLW00NS19,DBLP:journals/corr/abs-2202-01273}. These methods either disregard specific information about label noise or require strong statistical assumptions and tend to underperform under heavy noise and a large number of classes. Although these methods are theoretically grounded, their performance largely lags behind the fully-supervised counterpart.

\begin{figure}[t]
     \centering
     \begin{subfigure}{0.49\linewidth}
         \includegraphics[width=\columnwidth]{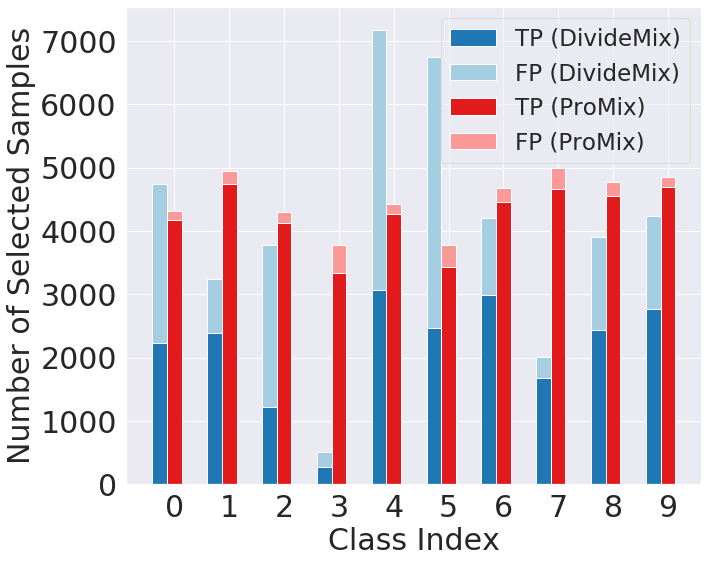}
         \caption{Number of selected samples.}
         \label{fig:headfigure_1}
     \end{subfigure}
     \hfill
     \begin{subfigure}{0.49\linewidth}
         \includegraphics[width=\columnwidth]{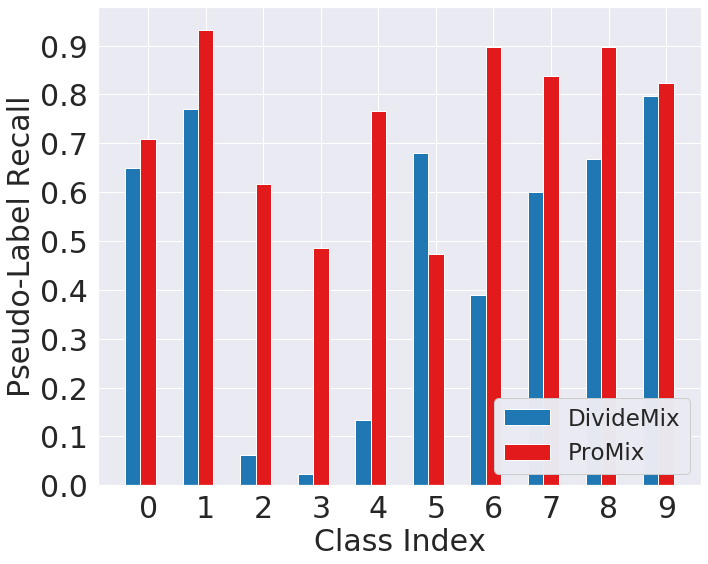}
         \caption{Recall of pseudo-labels.}
         \label{fig:headfigure_2}
     \end{subfigure}
     \caption{ (a) Comparison of selected samples for each class at the end of training on CIFAR-10 dataset with 90\% symmetric label noise. TP/FP 
 indicates true/false positive. (b) Comparison of pseudo-label recall for unchosen samples for each class at the end of training on CIFAR-10 dataset with 90\% symmetric label noise. }
 \vskip -0.1in
     \label{fig:headfigure}
\end{figure}

Another active line of research \cite{DBLP:conf/cvpr/WeiFC020,DBLP:conf/iclr/LiSH20,DBLP:conf/cvpr/YaoSZS00T21} attempts to filter out clean samples for training. For example, the pioneering Co-Teaching approach \cite{DBLP:conf/nips/HanYYNXHTS18} figures out that clean samples mostly pose smaller loss values due to the memorization effect of DNNs \cite{DBLP:conf/iclr/ZhangBHRV17}. Later, DivideMix \cite{DBLP:conf/iclr/LiSH20} further regards the remaining noisy samples as unlabeled and integrates well-developed semi-supervised learning algorithms. To date, this learning paradigm has been the de-facto standard for the LNL task \cite{DBLP:conf/cvpr/YaoSZS00T21,DBLP:conf/cvpr/KarimRRMS22,DBLP:journals/pr/CordeiroSBRC23} and shown to excel compared with early approaches. Despite the tremendous success, current selection-based methods solely ask for a decent-enough clean set with a restricted budget to ensure the high-precision of selected samples. 
By then, an abundance of valuable clean examples gets overlooked, while they possess great potential for performance enhancement.

Motivated by this, we present a novel framework ProMix which comprehensively investigates the opportunities and challenges of maximizing the utility of clean samples in LNL. To achieve this, we first propose a progressive selection mechanism that dynamically expands a base class-wise selection set by collecting those having high prediction scores on the observed labels. Empirically, such a simple modification allows us to reach an extraordinary trade-off between the quality and the quantity of the clean set. As shown in Figure~\ref{fig:headfigure_1}, compared with DivideMix \cite{DBLP:conf/iclr/LiSH20}, our progressive selection strategy selects much more clean samples while rarely introducing detested noise samples. Moreover, a simple baseline that is trained merely on such selected clean samples establishes the state-of-the-art performance on the real-world noisy CIFAR-N \cite{DBLP:conf/iclr/WeiZ0L0022} benchmark, and incorporating semi-supervised learning can further boost the performance. These results clearly validate the importance of well-utilizing clean samples. 

Despite the promising results, this maximal selection procedure inevitably causes side effects for the SSL as well. First, the clean samples may not be uniformly distributed across different labels, leading to a biased label distribution. Second, the selection and joint semi-supervised learning procedures are tightly interdependent, which may suffer from the confirmation bias stemming from its own hard-to-correct errors. To mitigate these problems, we develop a debiased semi-supervised training framework that comprises two key components. The first one is an auxiliary pseudo head that decouples the generation and utilization of unreliable pseudo-labels so as to alleviate the model from confirming its error recursively. Second, we incorporate a calibration algorithm that simultaneously rectifies the skewed pseudo-labels and the biased cross-entropy loss. As depicted in Figure~\ref{fig:headfigure_2}, integrated with debiased SSL, ProMix is able to generate pseudo-labels with higher quality on the unselected samples, thus facilitating the model training. By integrating the two key components into a whole framework, our proposed ProMix framework establishes state-of-the-art performance over various LNL benchmarks, even in the \emph{imbalanced LNL scenario} (see Appendix \ref{sec:app_b}). For example, on the CIFAR-10N datasets with real-world noisy labels, ProMix outperforms the best baseline by 2.04\%, 2.11\%, and 3.10\% under Aggregate, Rand1, and Worst label annotations respectively.

\section{Related Work}

\paragraph{Learning with Noisy Labels.}
Existing methods for LNL can be roughly categorized into two types. Earlier LNL approaches focus on loss correction. Since the vanilla cross-entropy (CE) loss has been proved to easily overfit corrupted data \cite{DBLP:conf/iclr/ZhangBHRV17}, a vast amount of robust losses has been designed to achieve a small and unbiased risk for unobservable clean data, including  robust variants of the CE loss \cite{DBLP:conf/iccv/0001MCLY019,DBLP:conf/nips/XuCKW19,DBLP:conf/icml/MaH00E020} or sample reweighted losses \cite{DBLP:conf/nips/ShuXY0ZXM19,DBLP:conf/aaai/ZhengAD21}. Some other methods \cite{DBLP:conf/cvpr/PatriniRMNQ17,DBLP:conf/nips/XiaLW00NS19,DBLP:journals/corr/abs-2202-01273} consider the noise transition matrix and depend on the estimated noise transition matrix to explicitly correct the loss function, but the transition matrix could be hard to estimate in the presence of heavy noise and a large number of classes. 

The latter strand of LNL methods tries to filter out a clean subset for robust training \cite{DBLP:conf/iclr/LiSH20,DBLP:conf/cvpr/KarimRRMS22,DBLP:journals/pr/CordeiroSBRC23}. Early methods \cite{DBLP:conf/nips/HanYYNXHTS18,DBLP:conf/icml/JiangZLLF18,DBLP:conf/cvpr/WeiFC020} typically adopt a peer network and perform small loss selection that is motivated by the memorization effect of DNNs \cite{DBLP:conf/iclr/ZhangBHRV17}. The seminal work DivideMix \cite{DBLP:conf/iclr/LiSH20} further incorporates MixMatch \cite{DBLP:conf/nips/BerthelotCGPOR19} to leverage the unselected examples and obtain extraordinary performance. This paradigm is then followed by the most recent LNL approaches \cite{DBLP:conf/aistats/LiSO20,DBLP:conf/nips/LiuNRF20,DBLP:conf/nips/BaiYHYLMNL21,DBLP:conf/cvpr/KarimRRMS22}. Recent LNL works also incorporate contrastive learning methods \cite{DBLP:conf/cvpr/OrtegoAAOM21,DBLP:conf/cvpr/YaoSZS00T21}. Despite the promise, existing selection-based methods mostly select a mediocre clean set with medium size and restricted budget, which overlooks a flurry of clean samples and naturally exhibits a quality-quantity trade-off. A too-small budget leads to inadequate clean sample utility, while a too-large budget enrolls many unexpected noisy examples. To cope with this, our work targets to maximize the utility of clean samples to unleash their great potential.

\paragraph{Debiased Learning.}
Debiased learning has gained huge attention since the source dataset and algorithms can be naturally biased \cite{DBLP:conf/cvpr/KimKKKK19,DBLP:conf/nips/NamCALS20}. For example, in long-tailed learning, models can be easily biased towards dominant classes due to the distribution bias in source data \cite{DBLP:journals/corr/abs-2110-04596}. 
As for semi-supervised learning, besides the distribution bias in source data and pseudo-labels \cite{DBLP:conf/cvpr/WangWLY22}, confirmation bias caused by inappropriate training with unreliable pseudo-labels may jeopardize generalization performance \cite{https://doi.org/10.48550/arxiv.2202.07136}. 
Debiased learning has also been a prevailing topic discussed in some other research topics like contrastive sampling \cite{DBLP:conf/nips/ChuangRL0J20} and OOD detection \cite{DBLP:conf/aaai/MingYL22}. 
Despite the surge in other areas, the bias in LNL still remains underexplored. In this work, we attempt to alleviate the biases in our selection and pseudo-labeling procedures. We hope our work will inspire future works to integrate this important component in advanced LNL frameworks.

\begin{figure*}[t]
     \centering
     \begin{subfigure}[][][t]{0.621\linewidth}
         \includegraphics[width=\columnwidth]{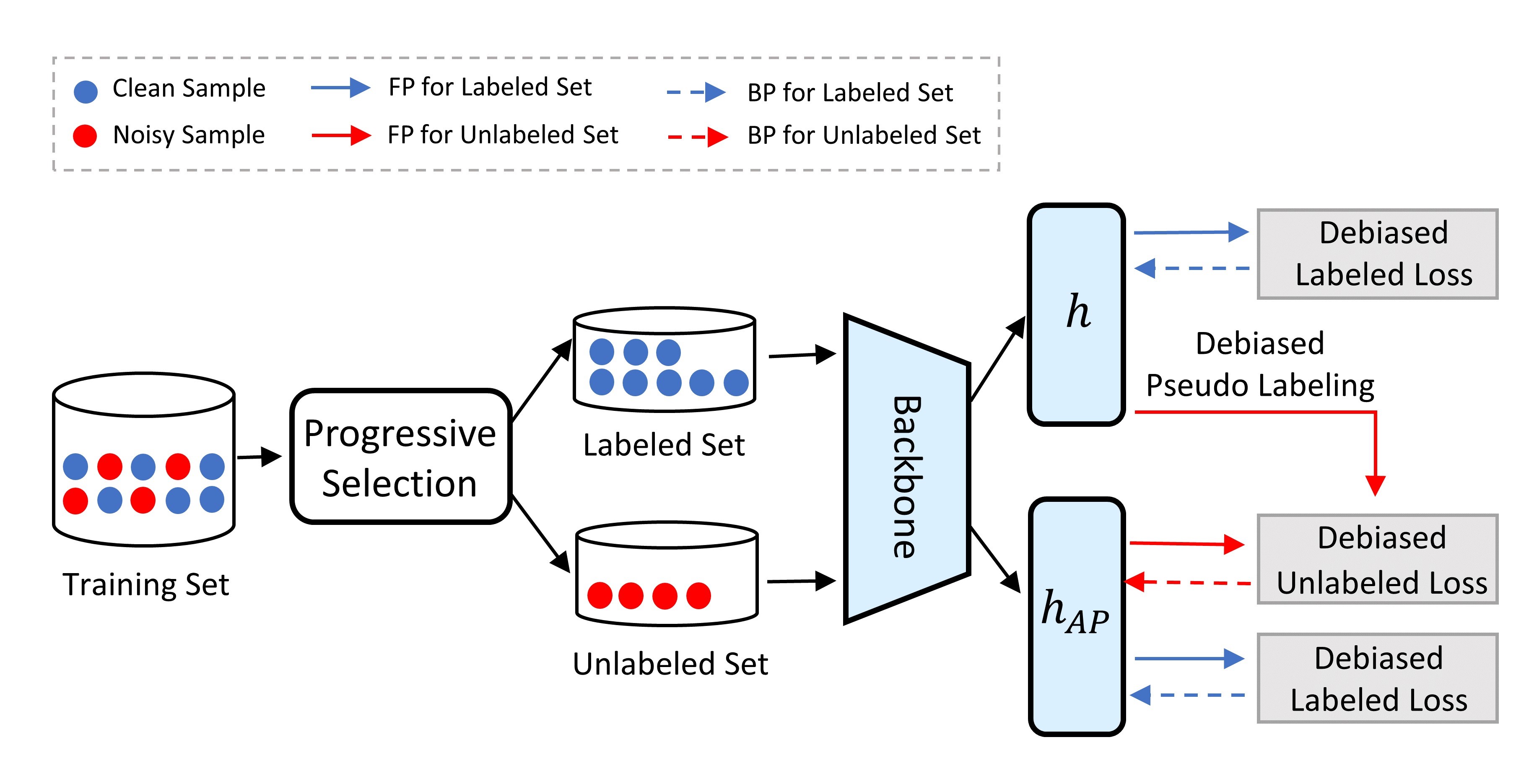}
         % \vspace{-1ex}
         \caption{Overall Framework of ProMix.}
     \end{subfigure}
     % \hfill
     \begin{subfigure}[][][t]{0.3\linewidth}
         \includegraphics[width=\columnwidth]{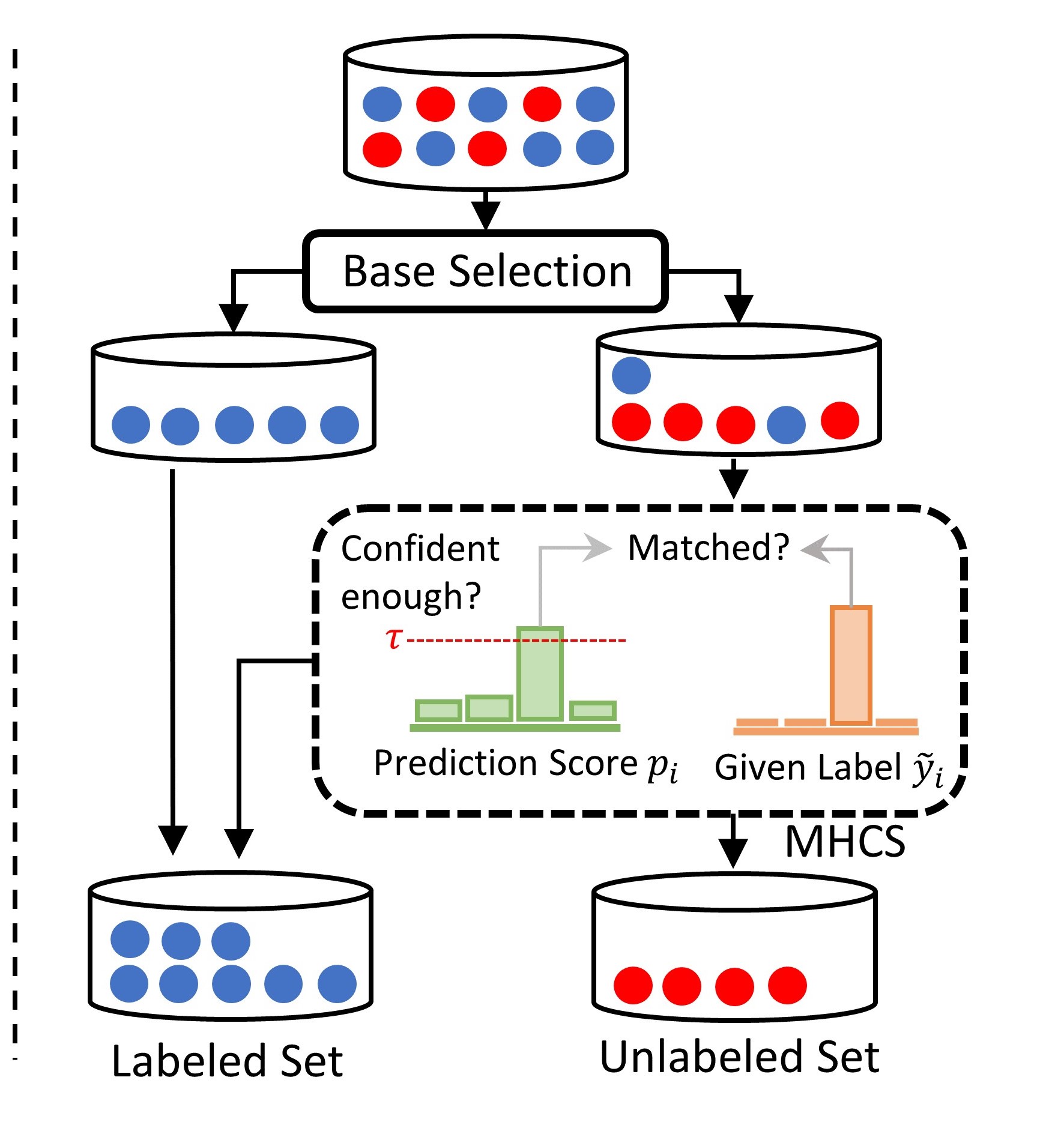}
         \caption{Progressive Selection.}
     \end{subfigure}
     \caption{\textbf{The left}: Overall framework of ProMix. After progressive selection, the source dataset is divided into the labeled set and unlabeled set for debiased SSL. The labeled data are fed forward to both primary head $h$ and auxiliary pseudo $h_{\text{AP}}$, while the unlabeled data are provided with debiased pseudo-labels by $h$ and then propagated forward through $h_{\text{AP}}$. Finally, the standard cross entropy in classification loss is replaced by debiased margin-based loss. FP/BP indicates forward/backward propagation. 
     \textbf{The right}: Details of progressive selection. The MHCS selects those examples with high confidence and matched predictions with given labels to dynamically expand a base selection set.
     }
     \label{fig:framework}
\end{figure*}

\section{Proposed Method}
\paragraph{Problem Formulation.}
For a multi-class classification task with $C$ classes, we denote the noisy training dataset with $n$ examples by $\mathcal{D}=\ \{(\bm{x}_i, {\tilde{{y}_i}})\}_{i=1}^n$, where each tuple comprises of an image $\bm{x}_i \in \mathcal{X} \subset \mathbb{R}^d $ and the corresponding \emph{noisy label}  ${\tilde{{y}_i}}\in \mathcal{Y} = \{1,...,C\}$. 
In contrast to the supervised learning setup, the ground-truth labels are not observable. 
The goal of LNL is to obtain a functional mapping $f_{\theta,\phi}:\mathcal{X}\mapsto \mathcal{Y}$ that takes an example $\bm{x}_i$ as the input and outputs the logits $f_{\theta,\phi}(\bm{x}_i)$ for predicting the true labels. This classification model $f_{\theta,\phi}= h_{\phi} \circ g_{\theta}$ consists of a feature extractor $g_{\theta}$ and a classification head $h_{\phi}$, which are parameterized by $\theta$ and $\phi$ respectively. In our work, we adopt the cross-entropy loss $\mathcal{H}=-\sum_{j=1}^{C}\tilde{{y}_i}^{j}\log{p_{i}^{j}}$
where $\bm{p}_i=\text{softmax}(f(\bm{x}_i))$ is the probabilistic output of $f(\bm{x}_i)$. $p_i^j$ is the $j$-th entry of $\bm{p}_i$ and $\tilde{{y}_i}^{j}$ is the $j$-th entry of one-hot encoded label vector ${\tilde{\bm{y}_i}}$. 

Similar to previous works, our framework also divides the samples into a clean set $\mathcal{D}_l$ and a noisy set $\mathcal{D}_u$, which are then applied for semi-supervised learning. The vanilla classification losses for these two parts are formulated:
\begin{equation}\label{formula:initial_loss}
\begin{split}
\mathcal{L}_x(\mathcal{D}_l) &= \frac{1}{|\mathcal{D}_l|}\sum\nolimits_{\bm{x}_i\in \mathcal{D}_l}\mathcal{H}({\tilde{\bm{y}_i}},\bm{p}_i)
\\
\mathcal{L}_u(\mathcal{D}_u) &= \frac{1}{|\mathcal{D}_u|}\sum\nolimits_{\bm{x}_i\in \mathcal{D}_u}\mathcal{H}(\text{Sharpen}(\bm{p}_i,T),\bm{p}_i)
\end{split}
\end{equation}
where $\text{Sharpen}(\bm{p}_i,T) =\frac{{(p_i^c)^\frac{1}{T}}}{{\sum_{c=1}^{C}}(p_i^c)^\frac{1}{T}} \text{ for }c=1,2,..,C$. An overview of the proposed method is illustrated in Figure~\ref{fig:framework}.

\subsection{Progressive Sample Selection}
As aforementioned, we would like to excessively collect clean samples. To achieve this, we first draw inspiration from the classical learning theory,
\begin{theorem}\cite{DBLP:books/daglib/0033642}
Let $\mathcal{F}$ be the hypothesis set of $f$. Denote the population risk by $R(f)=\mathbb{E}[\mathcal{H}]$ and the corresponding empirical risk by $\hat{R}(f)$. We define the true hypothesis by $f^*=\arg\min R(f)$ and empirical risk minimizer by $\hat{f}=\arg\min \hat{R}(f)$. Given a dataset containing $n$ examples, with probability at least $1-\delta$, the following inequality holds:
\begin{equation}
\begin{split}
R(\hat{f})\le R(f^*)+2\hat{\mathfrak{R}}(\mathcal{H})+5\sqrt{\frac{2\log(8/\delta)}{n}}
\end{split}
\end{equation}
where $\hat{\mathfrak{R}}(\mathcal{F})$ is the empirical Rademacher of $\mathcal{F}$.
\end{theorem}
Considering a simple case in which $f$ consists of multiple ReLU-activated fully-connected layers, we can apply the techniques outlined in \cite{DBLP:conf/icml/Allen-ZhuLS19} to derive an upper bound of the Rademacher complexity $O(\sqrt{\frac{1}{n}})$. If we assume a perfectly clean set with $n_c$ samples, the model has larger than a probability of $1-O(e^{-n_c\epsilon^2/50})$ to give an $\epsilon$-wrong prediction on the remaining dirty training data. Hence, this model is prone to produce a much larger probability on the observed labels for the remaining correctly-labeled data than noisy ones.
Motivated by this, we present the following selection procedure for enlarging the base selection set.

\paragraph{Class-wise Small-loss Selection (CSS).}
Based on the small loss criterion, we aim to first filter out a base clean  set as the cornerstone. In the training, the patterns of easy classes are usually fitted earlier and the loss values of examples with different observed labels may not be comparable \cite{DBLP:conf/ijcai/GuiWT21}. So it would be more appropriate to select examples class by class with the same observed labels. At each epoch, we first split the whole training dataset $\mathcal{D}$ to $C$ sets according to the noisy labels, i.e. $\mathcal{S}_j=\{(\bm{x}_i, {\tilde y}_i)\in \mathcal{D}|  {\tilde{y}_i}=j\}$. Given the $j$-th set $\mathcal{S}_j$, we calculate the standard CE loss $l_i$ for each example and select $k=\min(\lceil \frac{n}{C}\times R \rceil, |\mathcal{S}_j|)$ examples with smallest $l_i$ to constitute the clean set $\mathcal{C}_j$, where $R$ is the filter rate that is identical for all classes. Finally, the integral clean set is merged as $\mathcal{D}_{CSS} = \cup^C_{j=1} \mathcal{C}_j $. Compared to the original small-loss selection, we relate $k$ to the average number of examples $n/C$ such that produce a roughly balanced base set.

\paragraph{Matched High Confidence Selection (MHCS).}
After getting the base clean set, we also introduce another selection strategy to make capital out of the potentially clean samples missed by CSS. Specifically, we calculate the confidence scores $e_i=\max_j p^j_i$ for remaining samples. Then, we select those examples whose predictions match the given labels, while their confidence is high enough,
\begin{equation}\label{formula:mhcs}
\begin{split}
\mathcal{D}_{MHCS} = \{(\bm{x}_i, {\tilde y}_i)\in \mathcal{D}| e_i \ge \tau,{y'_i}={\tilde{y_i}} \}
\end{split}
\end{equation}
where ${y'_i}=\arg\max_{j}p^j_i$ is the predicted label of $\bm{x}_i$. In practice, we set a high threshold $\tau$ such that the selected samples have a high probability of being clean. According to our theoretical motivation, the proposed MHCS mechanism is able to automatically explore new clean data to dynamically enlarge the clean set $\mathcal{D}_{l} = \mathcal{D}_{CSS}\cup \mathcal{D}_{MHCS}$ by trusting those examples with matched high confidence.

\paragraph{Remark.} Our MHCS selection strategy draws inspiration from the pseudo-labeling procedure of FixMatch \cite{DBLP:conf/nips/SohnBCZZRCKL20}. But, in contrast to the SSL setup, our clean set is dynamically generated. With the naive high confidence selection, the DNNs can assign an arbitrary label with high confidence and select them as clean, which results in a vicious cycle. Thus, we choose those samples to have high confidence in their given labels, which tend to be originally clean.

    \subsection{Debiased Semi-Supervised Training}
After sample selection, we would also like to integrate semi-supervised learning that utilizes the remaining noisy samples to boost performance. Nevertheless, our excessive selection procedure inevitably induces unexpected characteristics. 
First, the selected clean samples may inherently be imbalanced amongst different labels, since some labels are typically more ambiguous than others. Hence, the model may be biased towards the classes that possess more clean samples and in turn pollutes the quality of selection (as shown in Figure~\ref{fig:headfigure_1}).
Second, the selection and pseudo-labeling are tightly coupled, which naturally brings about confirmation bias, so we should be particularly careful to train the model with the generated pseudo-labels. 
To escape such a dilemma, we introduce our debiased SSL framework in what follows.          
\begin{table*}[!t]
	\centering
 \small
	%\vspace{-1ex}
	% \renewcommand{\arraystretch}{1.1}% for the vertical padding	
	\tabcolsep=0.4cm
	\begin{tabular}{l || cccc | c||cccc} 
		\toprule	 	

			\multirow{2}{*}{Dataset } &\multicolumn{5}{c||}{CIFAR-10}&\multicolumn{4}{c}{CIFAR-100}\\
			\cmidrule{2-10}
			 &   \multicolumn{4}{c|}{Sym.}& Asym. & \multicolumn{4}{c}{Sym.}\\
			\midrule
			Methods\textbackslash Noise Ratio  & 20\% & 50\%& 80\% & 90\% &\multicolumn{1}{c||}{40\%}&20\% & 50\%& 80\% & 90\% \\
			\midrule
			
			{CE}	 &86.8&79.4	&62.9&42.7&85.0 & 62.0  & 46.7&19.9  &10.1 \\

			{Co-Teaching+}	 & 89.5 & 85.7 & 67.4 & 47.9 & - & 65.6  & 51.8 & 27.9  & 13.7 \\

            {JoCoR}	& 85.7& 79.4	&27.8& - &76.4 &53.0  &43.5& 15.5  & -  \\	             						 	
			
            {M-correction} & 94.0 & 92.0	&86.8&69.1&87.4 &73.9  &66.1 &48.2  &24.3 \\

            {PENCIL}	&92.4&89.1	&77.5&58.9&88.5 &69.4  &57.5&31.1  &15.3 \\

			{DivideMix}	&96.1&94.6	&93.2&76.0&93.4 & 77.3  & 74.6&60.2  &31.5 \\

			{ELR+}	&95.8 & 94.8	&93.3&78.7&93.0 &77.6  &73.6 &60.8  &33.4 \\	    
   
			{LongReMix}	&96.2&95.0	&93.9&82.0&94.7 &77.8  &75.6&62.9  &33.8 \\

			{MOIT}&94.1&91.1	&75.8&70.1& 93.2  &75.9  &70.1 &51.4  &24.5 \\

			{SOP+}	&96.3  & 95.5 &  94.0& - &93.8 &78.8  & 75.9 &  63.3  & - \\	   

			{PES(semi)}	&95.9&95.1	&93.1& - & -  &77.4 &74.3 &61.6  & -  \\

            {ULC}	&96.1&95.2	&94.0&86.4&94.6 &77.3  &74.9&61.2  &34.5 \\	             						 	
			\midrule

			\textbf{ProMix(last)}	 &\textbf{97.59}	&\textbf{97.30}&\textbf{95.05}&\textbf{91.13} &\textbf{96.51}&\textbf{82.39}  &\textbf{79.72}&\textbf{68.95}  &\textbf{42.74} \\	
   \textbf{ProMix(best)}	 &\textbf{97.69}	&\textbf{97.40}&\textbf{95.49}&\textbf{93.36} &\textbf{96.59}&\textbf{82.64}  &\textbf{80.06}&\textbf{69.37}  &\textbf{42.93} \\	
		\bottomrule
    \end{tabular}
	\caption{Accuracy comparisons on CIFAR-10/100 with symmetric (20\%-90\%) and asymmetric noise (40\%). We report both the averaged test accuracy over last 10 epochs and the best accuracy of ProMix. Results of previous methods are the best test accuracies cited from their original papers, where the blank ones indicate that the corresponding results are not provided. \textbf{Bold entries} indicate superior results.}
	\label{tab:cifar}
\end{table*}			

\paragraph{Mitigating Confirmation Bias. }

In most SSL pipelines, the generation and utilization of pseudo-labels are usually achieved by the same network blocks or between interdependent peer networks. 
As a result, the mistakes in pseudo-labels are nearly impossible to self-correct and can accumulate amplifying the existing confirmation bias. In order to alleviate this, we aim to disentangle the generation and utilization of pseudo-labels, which is achieved via introducing an extra classifier termed Auxiliary Pseudo Head (APH) $h_{\text{AP}}$. The primary classifier $h$ is the task-specific head that generates pseudo-labels for the unlabeled data but is merely optimized with labeled samples in $\mathcal{D}_l$. The newly devised classifier $h_{\text{AP}}$ share the same representations with $h$ but have independent parameters. It receives the pseudo-labels from $h$ and is trained with both labeled set $\mathcal{D}_l$ and unlabeled set $\mathcal{D}_u$. Based on Eq.~(\ref{formula:initial_loss}), the classification loss is formulated,
\begin{equation}\label{formula:classification_loss}
\begin{split}
\resizebox{0.9\hsize}{!}{
$\mathcal{L}_{cls}=\underbrace{\mathcal{L}_x(\{{\tilde{\bm{y}_i}},\bm{p}_i\})}_{\text {original head }}+\underbrace{\mathcal{L}_x(\{{\tilde{\bm{y}_i}},\bm{p}_i^{\prime}\}) + \lambda_u \mathcal{L}_u(\{{\text{Sharpen}(\bm{p}_i,T)},\bm{p}_i^{\prime}\})}_{\text {auxiliary pseudo head }}$
}
\end{split}
\end{equation}
where $\bm{p}_i$ is the softmax score from $h$ and $\bm{p}_i^{\prime}$ is that from $h_{\text{AP}}$. $\lambda_u$ controls the strength of the unsupervised loss. Through this procedure, the backbone benefits from the unlabeled data for enhanced representation via $h_{\text{AP}}$. At the same time, the primary classifier $h$ remains reliable without seeing unlabeled samples, which reduces the confirmation bias. It is worth noting that the auxiliary pseudo head only participates in training and is discarded at inference time.

\paragraph{Mitigating Distribution Bias. }
In addition to confirmation bias, there may still be distribution bias whose causes are two-fold. First, the selected samples can falsely calibrate the classifier due to skewed label distribution. Second, previous work \cite{DBLP:conf/cvpr/WangWLY22} shows that pseudo-labels can naturally bias towards some easy classes even if the training data are curated and balanced, which can iteratively dominate the training process. To learn a well-calibrated classifier, we incorporate a Debiased Margin-based Loss (DML) $l_{\text{DML}}$ to encourage a larger margin between the sample-rich classes and the sample-deficient classes. $l_{\text{DML}}$  can be defined as:
\begin{equation}
\begin{split}
l_{\text{DML}}=-\sum\nolimits_{j=1}^{C}\tilde{{y}_i}^{j}\log \frac{e^{f^{j}(\bm{x}_i)+\alpha \cdot \log \pi_{j}}}{\sum_{k=1}^C e^{f^k(\bm{x}_i)+\alpha \cdot \log \pi_{k}}}
\end{split}
\end{equation}
where $\alpha$ is a tuning parameter that controls debiasing strength and $\bm{\pi}$ is the underlying 
class distribution $\mathbb{P}(y \mid \bm{x})$ indicating which label possesses more samples on the training set. Here we assume $\bm{\pi}$ is known and will show how to estimate it later. The standard cross entropy loss $\mathcal{H}$ is then replaced by $l_{\text{DML}}$. 

Similarly, we also resort to Debiased Pseudo Labeling (DPL) following \cite{DBLP:conf/iclr/MenonJRJVK21} to calibrate the pseudo-labels. The refined logit $\tilde{f}_i$ which is then fed into the softmax layer for subsequent pseudo-label generation is formulated:
\begin{equation}\label{formula:dpl}
\begin{split}
\tilde{f}_i=f\left(\bm{x}_i\right)-\alpha \log \bm{\pi}
\end{split}
\end{equation}
That is, we suppress the logits on those easy classes and ensure other classes are fairly learned. The refined pseudo-labels are then used for calculating unlabeled loss in Eq.~(\ref{formula:classification_loss}). 

Next, we elaborate on our dynamical estimation of $\bm{\pi}$. Notably, we employ two different class distribution vectors for the labeled and unlabeled data respectively. 
Formally, given a mini-batch of samples $\mathcal{B}$ from either $\mathcal{D}_l$ or $\mathcal{D}_u$, we update the estimation of class distribution $\bm{\pi}$ in a moving-average style:
\begin{equation}
\begin{split}
\bm{\pi} = m \bm{\pi}+(1-m) \frac{1}{|\mathcal{B}|} \sum\nolimits_{\bm{x}_i \in \mathcal{B}} \bm{p}_i \\
\end{split}
\end{equation}
where $m \in [0,1)$ is a momentum coefficient. When calculating the loss in Eq.~(\ref{formula:classification_loss}), we use different estimation $\bm{\pi}_l$ and $\bm{\pi}_{u}$ for the two sets. For better understanding of the whole procedure, we refer the readers to Appendix \ref{sec:app_a} for more details. 

To summarize the pipeline of debiased SSL, the labeled data are fed forward to both $h$ and $h_{\text{AP}}$ while the unlabeled data are provided with the debiased pseudo-labels by $h$ as Eq.~(\ref{formula:dpl}) and then propagated forward through $h_{\text{AP}}$. Finally, the standard cross entropy in Eq.~(\ref{formula:classification_loss}) is replaced by $l_{\text{DML}}$. Through the above debiasing procedure, the refined pseudo-labels with higher quality and enhanced tolerance for incorrect pseudo-labels assist in milking unlabeled samples and in return enhance the purity of sample selection.

\subsection{Practical Implementation}
\paragraph{Label Guessing by Agreement (LGA). } While we still anticipate selecting matched high confidence instances, there are no free labels to be matched on falsely-labeled data whose predictions mismatch the observed labels. To further distill potential ``clean" samples from unlabeled data, we perform label guessing to correct noisy labels on the unlabeled set $\mathcal{D}_u$. We train two peer networks $f_{1}$ and $f_{2}$ in parallel to revise the noisy labels by agreement. Then we select a new subset $\mathcal{D}_c$ to enlarge the clean set $\mathcal{D}_l=\mathcal{D}_l \cup \mathcal{D}_c $ by the condition,
\begin{equation}
\begin{split}
(\max_jf_1^{j}(\bm{x})&\ge\tau)\wedge(\max_jf_2^j(\bm{x})\ge\tau)\wedge \\ &(\arg\max_jf^j_1(\bm{x})==\arg\max_jf^j_2(\bm{x}))
\end{split}
\end{equation}
In other words, the two networks should both hold high confidence on the same label. After that, we take the shared prediction $y'=\arg\max_jf_1^j(\bm{x})$ as the guessed labels for $\mathcal{D}_c $. With such reliable refurbishment, more purified clean samples could be involved in $\mathcal{D}_l$ to facilitate the SSL training.
\paragraph{Training Objective. }
To improve the representation ability of ProMix, we involve consistency training along with mixup augmentation on the reliable labeled set
$\mathcal{D}_l$. See Appendix \ref{sec:app_a} for more details. The losses for consistency regularization and mixup augmentation are denoted as $\mathcal{L}_{cr}$ and $\mathcal{L}_{mix}$ respectively. The overall training loss is given by,
\begin{equation}
\begin{split}
\mathcal{L}_{total}&=\mathcal{L}_{cls}+\gamma(\mathcal{L}_{cr}+\mathcal{L}_{mix})
\end{split}
\end{equation}
where $\gamma$ is a weighting factor. Notably, we ensemble the outputs of the two peer networks at the inference time.

\begin{table*}[!t]
 \centering

 \small
\begin{tabular}{l|cccccccccc|c}
\toprule
Method & CE    &  PENCIL & JoCoR    & DivideMix  & ELR+ & LongReMix  &  CORES*  & SOP & PES    & ULC & \textbf{ProMix} \\ \midrule
Accuracy & 69.21 & 73.49           & 70.30           & 74.76      & 74.81        & 74.38   &  73.24   & 73.50 & 74.64 & 74.90& \textbf{74.94} \\ \bottomrule
\end{tabular}
\caption{Accuracy comparisons on Clothing1M dataset. \textbf{Bold entries} indicate superior results.}
 \label{tab:clothing1m}
\end{table*}

\begin{table}[!t]
    \centering
    \small
    \begin{tabular}{l|ccc|c}
        \toprule
        \multirow{2}{*}{Methods}  & \multicolumn{3}{c|}{CIFAR-10N} &
        \multicolumn{1}{c}{CIFAR-100N} \\
        \cmidrule{2-5}
        & Aggre & Rand1  & Worst & Noisy Fine\\
        \midrule
        CE      & 87.77	  & 85.02  & 77.69  & 55.50\\ 
        Co-Teaching+    & 91.20	  & 90.33  & 83.83  & 60.37\\ 
        JoCoR   & 91.44	  & 90.30  & 83.37  & 59.97\\ 
        DivideMix   & 95.01	  & 95.16  & 92.56  & 71.13\\ 
        ELR+    & 94.83	  & 94.43  & 91.09  & 66.72\\ 
        CORES*  & 95.25	  & 94.45  & 91.66  & 55.72\\ 
        SOP+     & 95.61	  & 95.28  & 93.24  & 67.81\\ 
        PES(Semi)     & 94.66	  & 95.06  & 92.68  & 70.36\\ \midrule
        \textbf{ProMix}    & \textbf{97.65}	  & \textbf{97.39}  & \textbf{96.34}  & \textbf{73.79}\\
        \bottomrule
    \end{tabular}
    \caption{Accuracy comparisons on CIFAR-10N and CIFAR-100N under different noise types. \textbf{Bold entries} indicate superior results. }
    \label{tab:cifarn}
\end{table}

\section{Experiments}
In this section, we present the main results and part of the ablation results to demonstrate the effectiveness of the ProMix framework. We put more experimental results, including results on imbalanced dataset with synthetic noise, results on dataset with instance-dependent label noise and more results for ablation experiment in Appendix \ref{sec:app_b}.
\subsection{Setup}
\paragraph{Datasets.}
We first evaluate the performance of ProMix on CIFAR-10, CIFAR-100 \cite{krizhevsky2009learning}, Clothing1M \cite{DBLP:conf/cvpr/XiaoXYHW15} and ANIMAL-10N \cite{DBLP:conf/icml/SongK019} dataset. For CIFAR-10/100, we conduct experiments with synthetic symmetric and asymmetric label noise following the previous protocol \cite{DBLP:conf/cvpr/TanakaIYA18}. Symmetric noise is introduced by randomly flipping each ground-truth label to all possible labels with probability $r$ and asymmetric noise mimics structure of realistic noise taking place in similar classes. We also verify the effectiveness of ProMix on the recently proposed real-world noisy CIFAR-N \cite{DBLP:conf/iclr/WeiZ0L0022} dataset. The CIFAR-N dataset is composed of CIFAR-10N and CIFAR-100N, which equips the training samples of CIFAR-10 and CIFAR100 with human-annotated real-world noisy labels collected from Amazon Mechanical Turk. We adopt three noise types, Aggregate, Rand1 and Worst with a noise rate of 9.03\%, 17.23\% and 40.21\% respectively for CIFAR-10N and one noise type Noisy-Fine whose noise rate is 40.20\% for CIFAR-100N. Clothing1M dataset is a large-scale real-world noisy dataset containing 1 million images crawled from online shopping websites whose labels are collected by extracting tags from the surrounding texts. ANIMAL-10N is also a real-world noisy dataset consisting of human-annotated online images of ten confusable animals.

\paragraph{Implementation details.}
For CIFAR experiments, we use two ResNet-18 as the backbone of the peer networks. These networks are trained for 600 epochs, with a warm-up period of 10 epochs for CIFAR-10 and 30 epochs for CIFAR-100. We employ SGD as the optimizer with a momentum of 0.9 and weight decay of $5e^{-4}$. The batch size is fixed as 256 and the initial learning rate is 0.05, which decays by a cosine scheduler. The threshold $\tau$ is set as 0.99 and 0.95 for CIFAR-10 and CIFAR-100. The debiasing factor $\alpha$ is set as 0.8 and 0.5 for CIFAR-10 and CIFAR-100. $m$ is fixed as 0.9999. We leverage RandAugment to generate a strongly augmented view for consistency training.  The filter rate $R$ is set as $\{0.7,0.5,0.2,0.1\}$ for symmetric noise with noise ratio $r=\{0.2,0.5,0.8,0.9\}$ respectively and $0.5$ for asymmetric noise and all settings on CIFAR-N. LGA is performed at the last $K=250$ epochs such that the networks are accurate enough. We also set a maximum ratio threshold of total selected samples $\rho=0.9$ for all settings to restrict excessive correction. We linearly ramp up $\gamma$ from 0 to 1 and $\lambda_u$ from 0 to 0.1 to avoid overfitting false labels at the beginning. 

For Clothong1M, following previous work \cite{DBLP:conf/iclr/LiSH20}, the backbone is ResNet-50 pre-trained on ImageNet and the training is conducted for 80 epochs. We set learning rate as 0.002 and weight decay at 0.001. The filter rate $R$ is set as 0.7, $K$ is fixed as 40 and $\alpha$ is set at 0.8. We leverage AutoAugment \cite{DBLP:conf/cvpr/CubukZMVL19} with ImageNet-Policy for augmentation. For ANIMAL-10N, we adopt VGG-19 \cite{DBLP:journals/corr/SimonyanZ14a} with batch-normalization as backbone following \cite{DBLP:conf/icml/SongK019}. We use SGD with the same settings as CIFAR for 300 epochs of training. The filter rate $R$ is set as 0.7. {See Appendix \ref{sec:app_a} for more details.}

\subsection{Main Results}
We compare ProMix with multiple state-of-the-art methods. Specifically, they are Co-Teaching+ \cite{DBLP:conf/icml/Yu0YNTS19}, M-correction \cite{DBLP:conf/icml/ArazoOAOM19}, PENCIL \cite{DBLP:conf/cvpr/YiW19}, SELFIE \cite{DBLP:conf/icml/SongK019}, JoCoR \cite{DBLP:conf/cvpr/WeiFC020}, DivideMix \cite{DBLP:conf/iclr/LiSH20}, MOIT \cite{DBLP:conf/cvpr/OrtegoAAOM21}, PLC \cite{DBLP:conf/iclr/ZhangZW0021}, NCT \cite{DBLP:conf/cvpr/ChenSHS21}, ELR+ \cite{DBLP:conf/nips/LiuNRF20}, JoCoR \cite{DBLP:conf/cvpr/WeiFC020}, CORES* \cite{DBLP:conf/iclr/ChengZLGSL21}, PES \cite{DBLP:conf/nips/BaiYHYLMNL21}, SOP+ \cite{DBLP:journals/corr/abs-2202-14026}, SELC \cite{DBLP:conf/ijcai/LuH22}, ULC \cite{DBLP:conf/aaai/HuangBZBW22}, LongReMix \cite{DBLP:journals/pr/CordeiroSBRC23} and the naive cross-entropy (CE) method. For their performance, we directly adopt the reported results from their receptive papers. For fair comparisons, we follow the standard LNL evaluation protocols \cite{DBLP:conf/iclr/LiSH20,DBLP:conf/nips/XiaLW00NS19} and adopt the same backbone as in previous works for all benchmarks. 
\paragraph{Comparison with synthetic noisy labels. } 
 Table \ref{tab:cifar} shows the results on CIFAR-10/100 dataset with different types and levels of synthetic label noise. ProMix significantly outperforms all the rivals by a notable margin with different settings. Specifically, on the CIFAR-10 dataset with symmetric noise, ProMix exceeds the best baseline by 1.39\%, 1.90\%, 1.49\% and 6.96\% under the noise ratio of 20\%, 50\%, 80\%, 90\% respectively. Moreover, most baselines exhibit significant performance degradation under heavier noise, whereas ProMix displays robustness and remains competitive. These observations clearly verify the effectiveness of ProMix. 

\paragraph{Comparison with real-world noisy labels.}
As shown in Table \ref{tab:cifarn}, on the CIFAR-N dataset with real-world noisy labels, our proposed ProMix achieves the best performance compared with other existing methods on all four settings. In specific, on the CIFAR-10N dataset, the absolute accuracy gains for ProMix are 2.04\%, 2.11\% and 3.10\%  under Aggregate, Rand1, and Worst label annotations. Additionally, ProMix further expands the lead and reach a performance gain at 2.66\% on the even more challenging CIFAR-100N dataset. Table~\ref{tab:clothing1m} and Table~\ref{tab:animal} show that ProMix surpasses all the previous methods on the large-scale Clothing1M and ANIMAL-10N dataset, particularly achieves a lead of 5.88\% on ANIMAL-10N. These results clarify that our method can be well applied in real-world scenarios with large-scale data.

\begin{table}[t]
\centering
\small
\tabcolsep=0.18cm
\begin{tabular}{l|ccccc|c}
\toprule
Method& CE  & SELFIE  & PLC  & NCT  & SELC & \textbf{ProMix}      \\ \midrule
Accuracy & 79.4 & 81.8      & 83.4     & 84.1 & 83.7 & \textbf{89.98} \\ \bottomrule
\end{tabular}
\caption{Accuracy comparisons on ANIMAL-10N dataset. }
\label{tab:animal}
\end{table}

\begin{table*}[!ht]
    \centering
    \small
    \tabcolsep=0.35cm
    \begin{tabular}{l|cc|cccc}
        \toprule
        \multirow{2}{*}{Ablation}  & \multirow{2}{*}{LGA}  & \multirow{2}{*}{Selection Strategy} &  CIFAR-10 & CIFAR-100  & CIFAR-10N & CIFAR-100N\\ 
        & & & Sym.80\% & Sym.80\% &Worst&Noisy Fine \\
        \midrule
        \textbf{ProMix}     & \checkmark	  & CSS+MHCS  & \textbf{95.05}  & \textbf{68.95} & \textbf{96.34}  & \textbf{73.79}\\
        w/o LGA & \xmark	  & CSS+MHCS  & 94.32  & 61.19  & 95.57  & 73.10\\ 
        w/o MHCS      & \xmark	  & CSS& 93.17	  & 60.35  & 95.11  & 70.40\\ 
        w/o Base Selection    & \xmark	  & MHCS& 39.59	  & 21.01  & 74.09  & 40.75\\ 
        \midrule
       
        w/o CBR   & \xmark	  & CSS+MHCS & 93.96	  & 60.72  & 95.26  & 72.61\\ 
        w/o DBR   & \xmark	  & CSS+MHCS  & 94.07	  & 60.91  & 95.07  & 72.51\\ \midrule
         with Only Clean   & \xmark	  & CSS+MHCS & 93.82	  & 60.35  & 95.18  & 72.06\\ 
        
        \bottomrule
    \end{tabular}
    \caption{ \small Ablation study of ProMix on CIFAR-10-Symmetric 80\%, CIFAR-100-Symmetric 80\%, CIFAR-10N-Worst and CIFAR-100N-Noisy. ProMix with Only Clean denotes training with merely selected samples. CBR/DBR indicates Confirmation/Distribution Bias Removal. 
    } 
    \label{tab:ablation}
\end{table*}

\subsection{Ablation Study}

\paragraph{Effect of sample selection.}
We first dismiss the label guessing in variant \emph{ProMix w/o LGA}, where we note that LGA could boost performance, especially under severe noise. Based on this variant, we further equip ProMix with different sample selection strategies:  1) \emph{ProMix w/o MHCS} which performs CSS and overlooks the remaining matched high confidence samples; 2) \emph{ProMix w/o Base Selection} which discards the base selection and only selects matched high confidence samples. From Table~\ref{tab:ablation}, we can observe that ProMix w/o MHCS largely underperforms ProMix, which verifies the efficacy of MHCS mechanism. ProMix w/o Base Selection cannot establish a competitive performance because of inadequate labeled samples. This also proves that a clean-enough base selection set is the foundation of further progressive selection and debiased SSL. Moreover, We test a variant \emph{ProMix with Only Clean} which merely employs the selected samples and disregards the unselected samples. This variant suffers from perceptible performance degradation, but still beats all the baselines on CIFAR-N by a large margin. Such result indicates that clean samples are the real treasures in LNL, well-utilizing which is the key to successful learning.

\begin{figure}[t]
     \centering 
     \begin{subfigure}{0.49\linewidth}
         \includegraphics[width=\columnwidth]{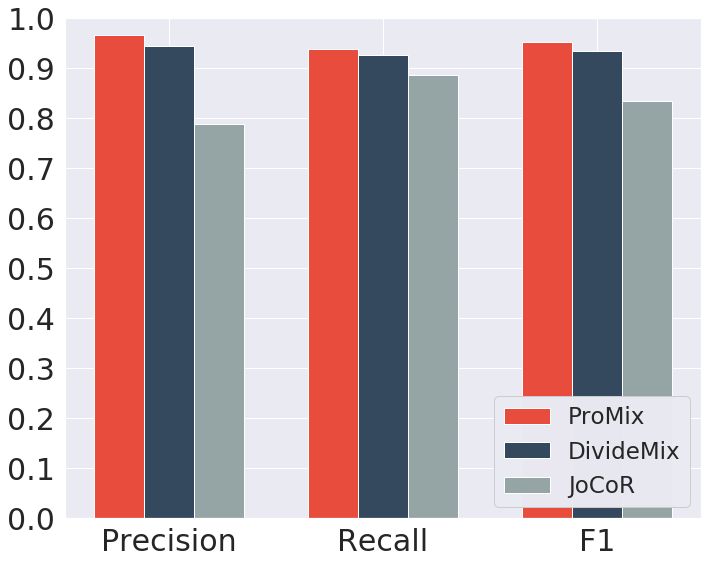}
         \caption{CIFAR-10-Symmetric 80\%.}
     \end{subfigure}
     \hfill
     \begin{subfigure}{0.49\linewidth}
         \includegraphics[width=\columnwidth]{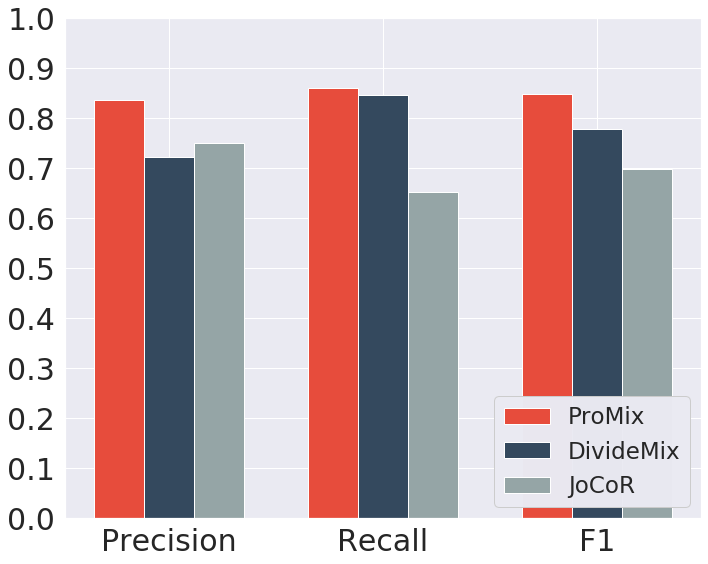}
         \caption{CIFAR-100-Symmetric 80\%.}
     \end{subfigure}
     \caption{Comparison of clean sample selection on CIFAR-10/100 dataset with 80\% symmetric noise. The threshold of DivideMix and forget rate of JoCoR have been re-tuned to get the best F1 score.}
     \label{fig:utility}
\end{figure}

\begin{figure}[t]
     \centering
     \begin{subfigure}{0.49\linewidth}
         \includegraphics[width=\columnwidth]{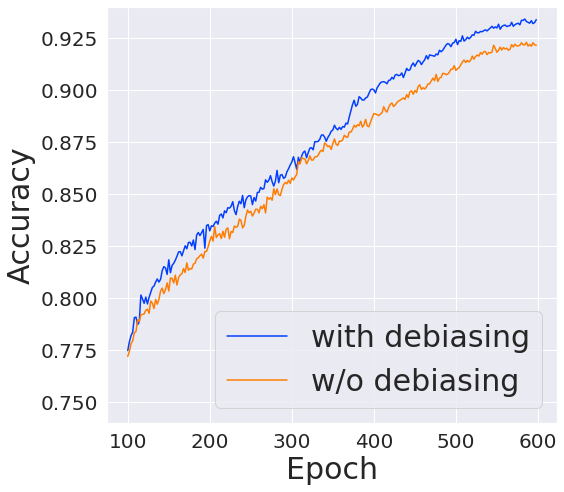}
         \caption{CIFAR-10-Symmetric 80\%.}
         \label{fig:cifar10pacc}
     \end{subfigure}
     \hfill
     \begin{subfigure}{0.49\linewidth}
         \includegraphics[width=\columnwidth]{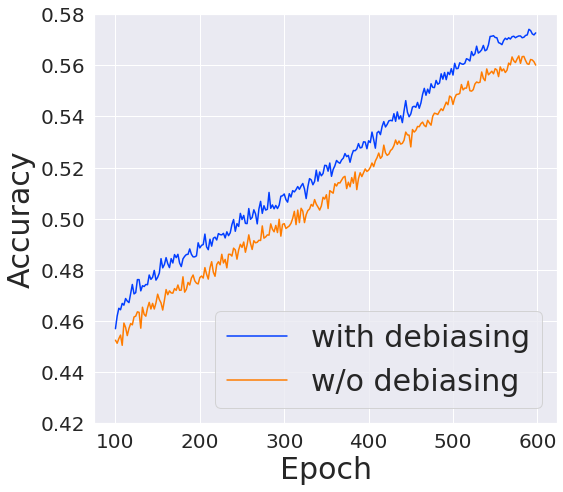}
         \caption{CIFAR-100-Symmetric 80\%.}
         \label{fig:cifar100pacc}
     \end{subfigure}
     \caption{Accuracy of pseudo-labels for the unlabeled data with and without debiasing on CIFAR-10/100 with 80\% Symmetric noise. LGA is disabled in order to avoid unexpected interference.  }
     \label{fig:pseudoacc}
\end{figure}

\paragraph{Effect of bias mitigation.}
Next, we ablate the contributions of the bias mitigation components in the debiased SSL. Specifically, we compare ProMix with two variants: 1) \emph{ProMix w/o Confirmation Bias Removal (CBR)} which discards $h_{\text{AP}}$, generating and utilizing the pseudo-labels on the same classification head; 2) \emph{ProMix w/o Distribution Bias Removal (DBR)} which sticks to the vanilla pseudo-labeling and adopts the standard cross entropy loss. From Table~\ref{tab:ablation}, we can observe that the lack of any one of these two components will bring about performance degradation, which indicates the effectiveness of bias mitigation. The performance drop grows more severe in imbalance scenarios as shown in Appendix \ref{sec:app_b}. Moreover, to reveal the impact of debiasing more intuitively, we compare the accuracy of the pseudo-labels of unlabeled data during training with and without incurring the bias mitigation strategies. As shown in Figure~\ref{fig:pseudoacc}, we can see that integrated with the debiasing strategies, the pseudo-labels on unlabeled data grow more accurate, hence facilitating the training process and reciprocating the sample selection process.

\paragraph{ProMix improves the clean sample utility.} To further prove that ProMix indeed improves the clean sample utility, we compare the precision and recall of filtered clean samples of ProMix and the other two baselines, DivideMix \cite{DBLP:conf/iclr/LiSH20} and JoCoR \cite{DBLP:conf/cvpr/WeiFC020}. From Figure~\ref{fig:utility}, we observe that ProMix significantly outperforms its competitors in both precision and recall, that is, ProMix selects much more clean samples while introducing fewer noisy samples. In specific, ProMix achieves a precision of 96.59\% and a recall of 93.78\% on CIFAR-10-Sym 80\%, with a lead of 2.23\% in precision and 1.15\% in recall compared to DivideMix. In more straightforward terms, ProMix picks 450 more clean samples and 891 fewer false positive noise samples than DivideMix, most of which are precious and meaningful hard samples, greatly facilitating training. Besides, on CIFAR-100-Sym 80\% where performances of competitors decline, ProMix leads by an even wider margin, especially in precision. This indicates that ProMix is based on a pipeline that first selects a clean-enough set with high precision, and then tries to expand the clean set with MHCS while still ensuring high precision, gradually maximizing clean sample utility. See Appendix \ref{sec:app_b} for more results of clean sample utility. 

\section{Conclusion}
In this work, we propose a novel framework called ProMix for learning with noisy labels with the principle of maximizing the utility of clean examples. ProMix dynamically expands a balanced base selection set by progressively collecting those samples with matched high confidence for the subsequent SSL. We also incorporate an auxiliary pseudo head and a calibration mechanism to mitigate the confirmation bias and the distribution bias in the process of training. The proposed framework is simple but achieves significantly superior performance to the state-of-the-art LNL techniques on different datasets with various kinds and levels of noise. We hope our work can inspire future research along this direction to promote the LNL methods by well utilizing the clean samples, which are the true treasures of the data. 
\section*{Acknowledgements}
Junbo Zhao wants to thank the support by the NSFC Grants (No. 62206247) and the Fundamental Research Funds for the Central Universities. Lei Feng is supported by the Joint NTU-WeBank Research Centre on Fintech (Award No: NWJ-2021-005), Nanyang Technological University, Singapore. This paper is also supported by Netease Youling Crowdsourcing Platform\footnote{https://fuxi.163.com}. The previous version won the \href{http://competition.noisylabels.com/}{1st Learning and Mining with Noisy Labels Challenge} in IJCAI-ECAI.

\appendix
\section{More Details of Implementation}\label{sec:app_a}
In this section, we describe more details of the practical implementation of ProMix.
\subsection{Details of Representation Enhancement}
\paragraph{Consistency Regularization.} Following previous works \cite{DBLP:conf/iclr/LiSH20}, we incorporate consistency regularization for boosted performance, which assumes that a classifier should produce similar predictions on a local neighbor of each data point. Given an image $\bm{x}_i$, we adopt two different data augmentation modules to obtain a weakly-augmented view $\bm{x}_i^w$ and
a strongly-augmented view $\bm{x_i^s}$, the CR loss $\mathcal{L}_{cr}$ has a similar form to the $\mathcal{L}_{cls}$, except that the source of $\bm{p}_i$ needs to be replaced from the weakly-augmented view  $\bm{x}_i^w$ to strongly-augmented view  $\bm{x}_i^s$. 

\paragraph{Mixup Augmentation.} We further incorporate mixup augmentation to encourage linear behavior between samples. A virtual mixed training example is generated by linearly interpolating
the randomly sampled pair of weakly-augmented examples $(\bm{x}_i^w,\tilde{\bm{y}_i})$ and $(\bm{x}_j^w,\tilde{\bm{y}_j})$
in $\mathcal{D}_l$ and taking a convex combination of labels as the
regression target. 
\begin{equation}
\begin{split}
\bm{x}^m & = \sigma {\bm{x}_i^w}+(1-\sigma) {\bm{x}_j^w} \\
\bm{y}^m & =\sigma \tilde{\bm{y}_i}+(1-\sigma) \tilde{\bm{y}_j}
\end{split}
\end{equation}
where $\sigma \sim \operatorname{Beta}(\varsigma, \varsigma)$ and we simply set $\varsigma = 4$ for CIFAR experiment and  $\varsigma = 0.5$ for Clothing1M experiment without further tuning. Besides interpolation-based data augmentation,
we also employ a masking-based mixed sample data
augmentation approach named FMix \cite{DBLP:journals/corr/abs-2002-12047} for exploiting local consistency.

\subsection{Details of Distribution Estimation}
Recall in the Distribution Bias Removal (DBR) component of ProMix, the underlying class distribution $\pi$ is estimated independently for the labeled set $\mathcal{D}_l$ and the unlabeled set $\mathcal{D}_u$. Notably, even if the original source dataset $\mathcal{D}$ is balanced, the separated two datasets can be both imbalanced. Considering a case where the clean samples are skewed distributed amongst labels, where some easy labels possess far more examples than some hard labels. Then, after selection, the clean set can exhibit a long-tailed distribution, whereas the dirty sample set can also be reversely skewed!

\begin{table}[h]
  
%   \vspace{5pt}

 \centering
 \small
	%\vspace{-1ex}
	% \renewcommand{\arraystretch}{1.1}% for the vertical padding	
	\tabcolsep=0.18cm

  \begin{tabular}{l|ccc|ccc}
    \toprule
    Dataset & \multicolumn{3}{c}{CIFAR-10IDN} & \multicolumn{3}{|c}{CIFAR-100IDN}\\
    \midrule
  
    Noise Ratio & 20\% & 40\% & 60\% & 20\% & 40\% & 60\% \\
    
    \midrule
    
    CE&85.45&76.23&59.75&57.79&41.15&25.28\\
    Forward \textit{T}&87.22&79.37&66.56&58.19&42.80&27.91\\
    $L_{\mathrm{DMI}}$&88.57&82.82&69.94&57.90&42.70&26.96\\
    Co-teaching&88.87&73.00&62.51&43.30&23.21&12.58\\
    Co-teaching+&89.90&73.78&59.22&41.71&24.45&12.58\\
    JoCoR&88.78&71.64&63.46&43.66&23.95&13.16\\
    Reweight-R&90.04&84.11&72.18&58.00&43.83&36.07\\
    Peer Loss&89.12&83.26&74.53&61.16&47.23&31.71\\
    CORES$^2$&91.14&83.67&77.68&66.47&58.99&38.55\\
    DivideMix&93.33&95.07&85.50&79.04&76.08&46.72\\
    CAL&92.01&84.96&79.82&69.11&63.17&43.58\\
    \midrule
    ProMix&\textbf{97.73}&\textbf{97.03}&\textbf{93.04}&\textbf{82.38}&\textbf{80.62}&\textbf{75.20}\\
    
    \bottomrule
  \end{tabular}

  \caption{Performance comparison for our method and the state-of-the-art methods on CIFAR-10IDN and CIFAR-100IDN.}\label{tab:idn}
\end{table}

\begin{table*}[t!]
\centering
 \small
	%\vspace{-1ex}
	% \renewcommand{\arraystretch}{1.1}% for the vertical padding	
	\tabcolsep=0.2cm
\begin{tabular}{ l c || c c c | c c c ||c c c|c c c}
\toprule

\multirow{2}{*}{{Dataset} } &&\multicolumn{6}{c||}{CIFAR-10}&\multicolumn{6}{c}{CIFAR-100}\\
			\cmidrule{3-14}
			 &  & \multicolumn{3}{c|}{$\text{Noise Ratio}=20\%$}& \multicolumn{3}{c||}{$\text{Noise Ratio}=50\%$} & \multicolumn{3}{c|}{$\text{Noise Ratio}=20\%$} & \multicolumn{3}{c}{$\text{Noise Ratio}=50\%$}\\
			\midrule
\multicolumn{2}{ l|| }{Methods\textbackslash Imbalance Factor} & 10 & 50 & 100 & 10 & 50 & 100 &  10 & 50 & 100 & 10 & 50 & 100  \\
\midrule
\multirow{2}{*}{CE}  & Best  & 77.86 & 64.38 & 61.79  & 60.72 & 46.50 & 38.43 & 45.97 & 33.41 & 29.85  & 28.70 & 18.49 & 16.24 \\
 & Last  & 74.00 & 61.38 & 55.69  & 44.29 & 32.69 & 27.78 & 45.75 & 33.12 & 29.58  & 23.70 & 16.56 & 14.19 \\
\midrule
\multirow{2}{*}{LDAM}  & Best  & 83.48 & 72.01 & 66.41  & 63.57 & 38.92 & 34.08 & 47.30 & 35.70 & 32.67  & 27.86 & 17.62 & 15.68  \\
 & Last  & 82.91 & 71.23 & 66.22  & 62.13 & 37.97 & 32.56 & 47.12 & 35.50 & 32.60  & 24.20 & 17.50 & 14.73 \\
\midrule
\multirow{2}{*}{LDAM-DRW}  & Best  & 84.98 & 76.77 & 73.24  & 69.53 & 49.90 & 42.60& 47.85 & 36.29 & 33.38  & 27.86 & 17.91 & 15.68  \\
 & Last  & 84.71 & 75.98 & 72.46  & 68.76 & 47.71 & 40.47 & 47.68 & 36.01 & 32.99  & 24.45 & 17.81 & 15.07  \\
\midrule
\multirow{2}{*}{DivideMix}  & Best  & 88.79 & 75.34 & 66.90  & 87.54 & 67.92 & 61.81 & 63.79 & 49.64 & 43.91  & 49.35 & 36.52 & 31.82  \\
 & Last  & 88.10 & 73.48 & 63.76  & 86.88 & 65.22 & 59.65 & 63.17 & 48.37 & 42.59  & 48.87 & 35.72 & 31.05 \\
\midrule
\multirow{2}{*}{RoLT+}  & Best  & 87.95 & 77.26 & 72.31  & 88.17 & 75.11 & 64.42 & 64.22 & 51.01 & 45.35  & 53.31 & 39.78 & 35.29  \\
 & Last  & 87.54 & 75.90 & 69.12  & 87.45 & 73.92 & 61.15  & 63.31 & 49.40 & 43.16  & 52.44 & 39.27 & 34.43 \\
\midrule
\multirow{2}{*}{Prototypical Classifier}  & Best  & 90.92 & 84.12 & 79.54 & 84.04 & 71.44 & 66.33 & 65.23 & 51.73 & 47.38 & 57.65 & 42.51 & 38.42\\
 & Last  & 90.81 & 83.71 & 78.34 & 83.51 & 71.44 & 64.69& 65.14 & 51.46 & 47.12 & 57.65 & \textbf{42.51} & \textbf{38.36}  \\
 \midrule
\multirow{2}{*}{ProMix}  & Best  & \textbf{94.40} & \textbf{86.05} & \textbf{81.63} & \textbf{91.83} & \textbf{76.14} & \textbf{68.79} & \textbf{67.11} & \textbf{51.99} & \textbf{48.67} & \textbf{60.27} & \textbf{42.96} & \textbf{38.52}\\
 & Last  & \textbf{94.22} & \textbf{85.12} & \textbf{79.69} & \textbf{91.71} & \textbf{73.04} & \textbf{67.43} & \textbf{66.95} & \textbf{51.59} & \textbf{48.57} & \textbf{59.99} & 42.28 & 38.21  \\
\bottomrule
\end{tabular}
\caption{ Accuracy comparisons on imbalanced CIFAR-10 and CIFAR-100 dataset. \textbf{Bold entries} indicate superior results. }\label{tab:imb}
\end{table*}

Now, we elaborate on our intuition of estimating two different class priors for $\mathcal{D}_l$ and $\mathcal{D}_u$. Assume that we are given an arbitrary dataset $\mathcal{D}$ and its prior distribution $\pi$, we typically perform logit calibration \cite{DBLP:conf/iclr/MenonJRJVK21} to find a balanced classifier. With a slight abuse of notations, we have,
\begin{equation}\label{formula:global_min}
\begin{split}
f^\text{balanced}= {\arg \min}_f \mathcal{L}(\mathcal{D},\pi)
\end{split}
\end{equation}
For our ProMix, our target is,
\begin{equation}\label{formula:target}
\begin{split}
\text{min} \mathcal{L}(\mathcal{D}_l,\pi_l)+\mathcal{L}(\mathcal{D}_u,\pi_u)
\end{split}
\end{equation}
where $\pi_l$ and $\pi_u$ are the prior label distribution of $\mathcal{D}_l$ and $\mathcal{D}_u$ respectively. According to Eq.~(\ref{formula:global_min}), we can observe that $f^\text{balanced}$ is the minimizer of both two terms, and hence, is also the global optimizer of Eq.~(\ref{formula:target}). Accordingly, our ProMix framework oughts to estimate two different distribution vectors for the two sets separately.

It is worth noting that we also perform the pseudo-label calibration. Theoretically, it is analogous to setting a larger margin for the DML loss $l_{\text{DML}}$. We believe a high-quality pseudo-label may further boost the performance. In effect, we empirically find that it does indeed works well. 

Lastly, we note that our debiasing procedure can be more effective in imbalanced LNL setups, as the distributional bias can be more distinguishable. We henceforth conducted more ablations and the results are listed in Appendix~\ref{imb:result}.

\subsection{Additional implementation details}
For Clothong1M, we adopt a class-wise Gaussian Mixture Model following \cite{DBLP:conf/iclr/LiSH20} to replace small-loss selection for filtering out a base set. For ANIMAL-10N, We use SGD with the same settings as CIFAR for 300 epochs of training. The filter rate $R$ is set as 0.7 and $K$ is fixed as 150.

\section{Additional Experimental Results}\label{sec:app_b}
In this section, we report additional empirical results of ProMix. We compare ProMix with multiple state-of-the-art methods, including some newly introduced baselines. Specifically, for comparisons on imbalanced noisy datasets, they are LDAM \cite{DBLP:conf/nips/CaoWGAM19}, LDAM-DRW \cite{DBLP:conf/nips/CaoWGAM19}, DivideMix \cite{DBLP:conf/iclr/LiSH20}, RoLT \cite{DBLP:journals/corr/abs-2108-11569}, Prototypical Classifier \cite{DBLP:conf/pakdd/WeiSLZ22}. For comparisons with instance-dependent noise and real-world noise, they are Reweight-R \cite{DBLP:conf/nips/XiaLW00NS19}, $L_{\mathrm{DMI}}$ \cite{DBLP:conf/nips/XuCKW19}, Forward $T$ \cite{DBLP:conf/cvpr/PatriniRMNQ17},  Peer Loss \cite{DBLP:conf/icml/LiuG20}, Positive-LS \cite{DBLP:conf/icml/LukasikBMK20}, Negative-LS \cite{DBLP:journals/corr/abs-2106-04149}, VolMinNet \cite{DBLP:conf/icml/LiL00S21}, CAL \cite{DBLP:conf/cvpr/ZhuL021}. All experiments are conducted on a workstation with 8 NVIDIA RTX A6000 GPUs. We run all experiments five times and report the mean results. 

\begin{table}[!t]
    \centering
    \small
    \tabcolsep=0.2cm
    \begin{tabular}{l|ccc|c}
        \toprule
        \multirow{2}{*}{Metrics}& \multicolumn{3}{c|}{CIFAR-10N} & CIFAR-100N \\
        \cmidrule{2-5}
        & Aggregate  & Rand1  & Worst & Noisy Fine   \\
        \midrule
        Precision      &  99.10    & 98.83  & 96.20 & 85.66\\
        Recall         &  99.01	& 98.82  & 98.71 & 89.82\\
        F1-score      & 99.05	& 98.82  & 97.44 & 87.69 \\
        \bottomrule
    \end{tabular}
    \caption{Results of clean sample utility of ProMix on CIFAR-10N. }
    \label{tab:ciarn_uti}
\end{table}

\begin{table*}[!t]
    \centering
    \small
    \tabcolsep=0.4cm
    \begin{tabular}{l|ccccc|c}
        \toprule
        \multirow{2}{*}{Methods}  & \multicolumn{5}{c|}{CIFAR-10N} &
        \multicolumn{1}{c}{CIFAR-100N} \\
        \cmidrule{2-7}
		 & Aggre & Rand1 & Rand2& Rand3 & Worst & Noisy Fine\\
        \midrule
			     CE   & 87.77 $\pm$ 0.38 &  85.02 $\pm$ 0.65& 86.46 $\pm$ 1.79  & 85.16 $\pm$ 0.61 & 77.69 $\pm$ 1.55 & 55.50 $\pm$ 0.66\\
				 Forward $T$   & 88.24 $\pm$ 0.22 & 86.88 $\pm$ 0.50 & 86.14 $\pm$ 0.24 & 87.04 $\pm$ 0.35 &  79.79 $\pm$ 0.46  & 57.01 $\pm$ 1.03\\
				 Co-teaching+  & 90.61 $\pm$ 0.22 & 89.70 $\pm$ 0.27 & 89.47 $\pm$ 0.18 & 89.54 $\pm$ 0.22 & 83.26 $\pm$ 0.17 & 57.88 $\pm$ 0.24\\
				Peer Loss   & 90.75 $\pm$ 0.25 & 89.06 $\pm$ 0.11 & 88.76 $\pm$ 0.19 & 88.57 $\pm$ 0.09 & 82.00 $\pm$ 0.60 &  57.59 $\pm$ 0.61\\
			ELR+   & 94.83 $\pm$ 0.10 & 94.43 $\pm$ 0.41 & 94.20 $\pm$ 0.24 & 94.34 $\pm$ 0.22 & 91.09 $\pm$ 1.60  & 66.72 $\pm$ 0.07 \\	
   Positive-LS  & 91.57 $\pm$ 0.07 & 89.80 $\pm$ 0.28 & 89.35 $\pm$ 0.33 & 89.82 $\pm$ 0.14 & 82.76 $\pm$ 0.53  & 55.84 $\pm$ 0.48\\
				F-Div   & 91.64 $\pm$ 0.34 & 89.70 $\pm$ 0.40 & 89.79 $\pm$ 0.12 & 89.55 $\pm$ 0.49 & 82.53 $\pm$ 0.52 & 57.10 $\pm$ 0.65\\
				Divide-Mix   & 95.01 $\pm$ 0.71  & 95.16 $\pm$ 0.19 & 95.23 $\pm$ 0.07  & 95.21 $\pm$ 0.14  & 92.56 $\pm$ 0.42  & 71.13 $\pm$ 0.48\\
				Negative-LS   & 91.97 $\pm$ 0.46 & 90.29 $\pm$ 0.32 & 90.37 $\pm$ 0.12 & 90.13 $\pm$ 0.19 & 82.99 $\pm$ 0.36 &  58.59 $\pm$ 0.98\\
				 CORES$^*$    &  95.25 $\pm$ 0.09 & 94.45 $\pm$ 0.14  &   94.88 $\pm$ 0.31& 94.74 $\pm$ 0.03 & 91.66 $\pm$ 0.09 &  55.72 $\pm$ 0.42\\
				  VolMinNet    & 89.70 $\pm$ 0.21  & 88.30 $\pm$ 0.12 & 88.27 $\pm$ 0.09  & 88.19 $\pm$ 0.41 & 80.53 $\pm$ 0.20  & 57.80 $\pm$ 0.31\\
				CAL   & 91.97 $\pm$ 0.32 & 90.93 $\pm$ 0.31 & 90.75 $\pm$ 0.30 & 90.74 $\pm$ 0.24 & 85.36 $\pm$ 0.16 &  61.73 $\pm$ 0.42\\
					PES (Semi)  &  94.66 $\pm$ 0.18    & 95.06 $\pm$ 0.15 & 95.19 $\pm$ 0.23 &  95.22 $\pm$ 0.13 & 92.68 $\pm$ 0.22  &  70.36 $\pm$ 0.33 \\ \midrule
        \textbf{ProMix}    & \textbf{97.65 $\pm$ 0.19}	  & \textbf{97.39 $\pm$ 0.16 } & \textbf{97.55 $\pm$ 0.12 }  & \textbf{97.52 $\pm$ 0.09}  & \textbf{96.34 $\pm$ 0.23}   &\textbf{73.79 $\pm$ 0.28}\\
        \bottomrule
    \end{tabular}
    \caption{Accuracy comparisons on CIFAR-10N and CIFAR-100N under different noise types. \textbf{Bold entries} indicate superior results. }
    \label{tab:cifarn_all}
\end{table*}
\begin{table}[!t]
    \centering
    \small
    \begin{tabular}{l|ccccc}
        \toprule
        Threshold $\tau$ & 0.9  & 0.92  & 0.95  & 0.99 & 0.999 \\
        \midrule
        CIFAR-10 & 95.20& 95.33& 95.58& 95.49& 95.45 \\
        CIFAR-100 & 69.31& 69.34& 69.37& 69.01& 68.69\\
        \bottomrule
    \end{tabular}
    \caption{Ablation study of different confidence threshold $\tau$.}
    \label{tab:ablation_tau}
\end{table}

\subsection{Comparisons on imbalanced noisy dataset}\label{imb:result}
To demonstrate the robustness of ProMix under skewed distribution in source data, We further conduct experiments on  imbalanced datasets with synthetic noise.
Formally, we generate the class-imbalanced dataset following previous works \cite{DBLP:conf/pakdd/WeiSLZ22}. For a dataset with $C$ classes and $N$ training examples for each class, by assuming the imbalance factor is $\kappa$, the number of examples for the $i$-th class is set to $N_i={N}/{\kappa^{\frac{i-1}{C-1}}}$. As the number of samples for each class in source data is heavily skewed, we update the number of samples for base selection in CSS from $k=\min(\lceil \frac{n}{C}\times R \rceil, |\mathcal{S}_j|)$ to $k= (|\mathcal{S}_j|\times R)$ for $j$-th subset $\mathcal{S}_j$.

As shown in Table~\ref{tab:imb}, ProMix outperforms all the rivals in most settings. On the imbalanced CIFAR-10 with 20\% label noise. ProMix leads with a margin of 3.58\%, 1.94\%, and 2.09\% with the imbalance factor of 10, 50, and 100 respectively. When the noise ratio gets higher, The leading margin for ProMix is further expanded, especially when the imbalance of data is not severe. These observations demonstrate the superiority and robustness of ProMix for the LNL task in the imbalanced scenario. 

Moreover, we also conduct ablation experiments of the bias mitigation components in such imbalanced scenarios to demonstrate their contributions more intuitively. As shown in Table~\ref{tab:ablation_imb}, the performance of \emph{ProMix w/o Distribution Bias Removal (DBR)} predominantly lags behind that of ProMix, especially when the imbalance factor grows larger. \emph{ProMix w/o Contribution Bias Removal (CBR)} also suffers from severe performance degradation. These results further verify the effectiveness of bias mitigation components, which is more significant in the imbalanced scenario. 

\begin{figure}[!t]
     \centering
     \begin{subfigure}{0.49\linewidth}
         \includegraphics[width=\columnwidth]{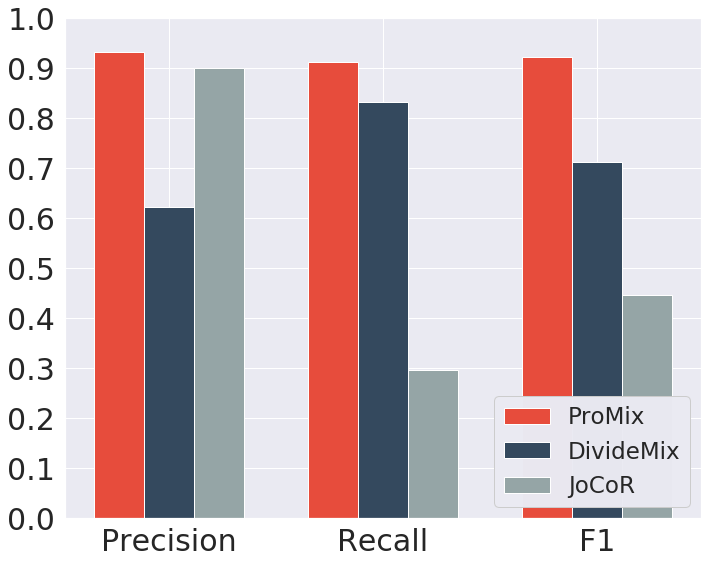}
         \caption{CIFAR-10 Symmetric 90\%.}
     \end{subfigure}
     \hfill
     \begin{subfigure}{0.49\linewidth}
         \includegraphics[width=\columnwidth]{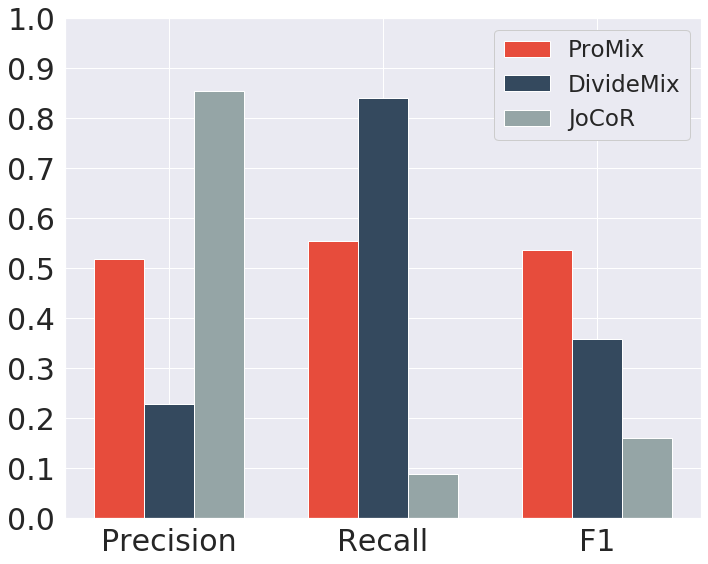}
         \caption{CIFAR-100 Symmetric 90\%.}
     \end{subfigure}
     \caption{Comparison of clean sample selection on CIFAR-10 and CIFAR-100 dataset with 90\% symmetric label noise. The threshold of DivideMix and the forget rate of JoCoR have been re-tuned to get the best results.}
     \label{fig:more_uti}
\end{figure}

\subsection{Comparisons with instance-dependent noise}
We also conduct experiments on datasets with instance-dependent label noise following the previous protocol \cite{DBLP:conf/cvpr/ZhuL021} on the CIFAR-10 and CIFAR-100 dataset with the noise ratio of 20\%, 40\% and 60\% respectively.

As shown in Table~\ref{tab:idn}, ProMix largely outperforms all the baselines on different datasets and noise ratios with a prominent lead. When the noise ratio increases to high and other methods mostly exhibit a severe performance drop, ProMix consistently remains competitive and shows robustness. Such results demonstrate that ProMix is solid under the condition of instance-dependent label noise.

\subsection{Complete results on CIFAR-N dataset}
We include a larger family of LNL methods and the noise types of CIFAR-10N-Rand2 and CIFAR-10N-Rand3 annotation to display the full results on the CIFAR-N dataset in Table~\ref{tab:cifarn_all}. We can observe that ProMix exceeds all the opponents by a large margin on all settings, including Rand2 and Rand3. This further verifies the effectiveness of ProMix on the real-world noisy CIFAR-N dataset.

\subsection{More Ablations}
\paragraph{More results of clean sample utility.}
We also provide more results of clean sample utility to clarify the strong distinguishing ability of ProMix.
As shown in Figure~\ref{fig:more_uti}, we can see that on CIFAR-10-Symmetric 90\%, ProMix vastly outperforms competitors in both precision and recall. On CIFAR-100-Symmetric 90\%, DivideMix pursues a high recall at the expense of extremely low precision. That is, DivideMix adopts a too-large budget and enrolls numerous noisy samples, while JoCoR similarly embraces a too-small budget instead. ProMix reaches an extraordinary trade-off between them and produces the higher F1-score. Additionally, we display the results for the utilization of clean samples on real-world noisy dataset CIFAR-N in Table~\ref{tab:ciarn_uti}, we can see that ProMix is able to achieve a precision of 99.10\% and a recall of 99.01\% on CIFAR-N-Aggregate, which indicates that after the progressive selection and label guessing by agreement, the training examples finally become near supervised ones. Based on these observations, We again conclude that ProMix does indeed maximize the clean sample utility. 

\begin{table}[!t]
    \centering
    \small
    \tabcolsep=0.15cm
    \begin{tabular}{l|ccc|ccc}
        \toprule
        {Dataset} & \multicolumn{3}{c|}{CIFAR-10}  & \multicolumn{3}{c}{CIFAR-100} \\ \midrule 
        Noise Ratio & \multicolumn{6}{c}{20\%} \\
        \midrule
        Imbalance Factor & 10 & 50 & 100 & 10 & 50 & 100\\
        \midrule
        \textbf{ProMix}  & \textbf{94.40}& \textbf{86.05} & \textbf{81.63} & \textbf{67.11} & \textbf{51.99} & \textbf{48.67} \\
        w/o DBR & 92.71 & 83.82 & 75.72 & 64.03& 47.95 & 43.09 \\
        w/o CBR & 93.39 & 83.69 & 77.18 & 65.27& 49.10 & 45.43 \\
        \bottomrule
    \end{tabular}
    \caption{Ablation study of bias mitigation on the imbalanced CIFAR-10 and CIFAR-100 with 20\% label noise. CBR/DBR indicates Confirmation/Distribution Bias Removal. }
    \label{tab:ablation_imb}
\end{table}

\begin{figure}[!t]
     \centering
         \includegraphics[width=0.66\columnwidth]{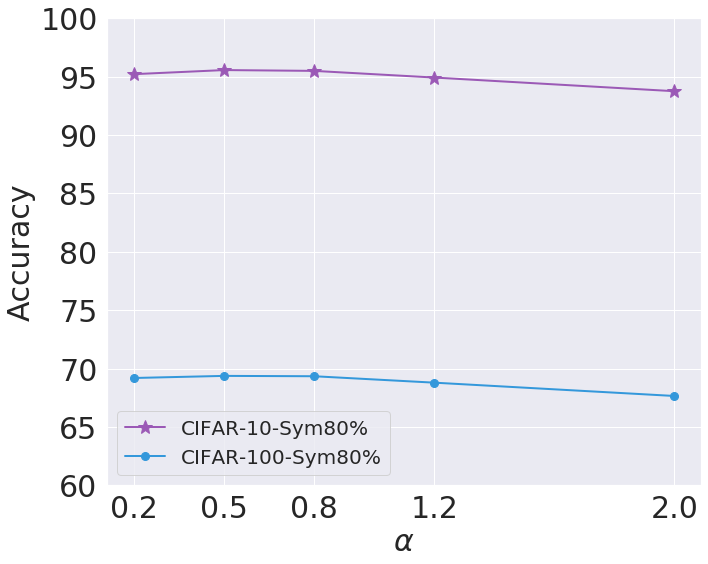}
     \caption{ Ablation study of different debiasing factor  $\alpha$. }
     \label{fig:ablation_alpha}
\end{figure}

\paragraph{Effect of confidence threshold $\tau$.}
We further investigate the effect of the high-confidence threshold $\tau$. Table~\ref{tab:ablation_tau} shows the performance of ProMix with different $\tau$ on CIFAR-10-Symmetric 80\% and CIFAR-100-Symmetric 80\%. It can be observed that ProMix works well in a wide range of $\tau$ values on CIFAR-10. While for CIFAR-100, ProMix bears perceptible performance drop when $\tau$ gets too high. The reason might be CIFAR-10 is more simple and the output probability can be confident enough quickly, while a too-high threshold could be hard to reach in a much larger label space.

\paragraph{Effect of debiasing factor $\alpha$.}
Then, we show the effect of $\alpha$ that controls the strength of debiasing in Figure~\ref{fig:ablation_alpha}. ProMix is able to produce a performance gain when $\alpha$ grows larger from 0.2 to 0.8 and 0.5 for CIFAR-10 and CIFAR-100 respectively. But as $\alpha$ continues to get too large, ProMix shows inferior results on both datasets. 
In practice, we empirically choose a moderate value without further careful fine-tuning.

\section{Overall Algorithm}
We describe the overall pipeline of ProMix in Algorithm \ref{alg:algorithm}. 

\begin{algorithm}[h!]
\small
    \DontPrintSemicolon
    \SetNoFillComment
    \caption{Training pipeline of ProMix.}
    \label{alg:algorithm}
    \textbf{Input:} Training set $\mathcal{D}=\ \{(\bm{x}_i, {\tilde{{y}_i}})\}_{i=1}^n$, the classification model $f_{\theta,\phi}= h_{\phi} \circ g_{\theta}$, primary head $h$, auxiliary pseudo head $h_{\text{AP}}$, debiasing parameter $\alpha$, loss weighting factors $\gamma, \lambda_u$ , the start epoch $K$ for LGA \\
    $\theta,\phi = \text{WarmUp}(\mathcal{D},\theta,\phi)$ \\
    \For {$epoch=1,2,\ldots,$}
    {
        \tcp{class-wise small-loss selection}
        \For {$j=1,2,\ldots,C$}{
        $\mathcal{S}_j=\{(\bm{x}_i, {\tilde y}_i)\in \mathcal{D}|  {\tilde{y}_i}=j\}$ \\
        Calculate each example's CE loss $l_i$ \\
        Select $k$ samples with smallest $l_i$ as $\mathcal{C}_j$  \\
        }
        $\mathcal{D}_{CSS} = \cup^C_{j=1} \mathcal{C}_j $ \\ 
        \tcp{matched high confidence selection}
        $\mathcal{D}_{MHCS} = \{(\bm{x}_i, {\tilde y}_i)\in \mathcal{D}| e_i \ge \tau,{y'_i}={\tilde{y_i}} \}$ \\ 
        $\mathcal{D}_{l} = \mathcal{D}_{CSS}\cup \mathcal{D}_{MHCS}$  \\
       \tcp{label guessing}
        \If{$epoch\ge K$}{
        Select high-confidence examples with \\
        agreed predictions as $\mathcal{D}_c$ according to Eq .(8) \\ 
       $\mathcal{D}_l=\mathcal{D}_l \cup \mathcal{D}_c $ \\
        }
        \tcp{debiased semi-supervised training}
        $\mathcal{D}_u = \mathcal{D} \setminus \mathcal{D}_l$, mini-batch $\mathcal{B}$ sampled from $\mathcal{D}_l$ or $\mathcal{D}_u$ \\
        $\mathcal{D}_l$ are fed forward to both $h$ and $h_{\text{AP}}$. \\
        $\mathcal{D}_u$ are given with debiased pseudo-label by $h$ as Eq. (6) \\
        $\mathcal{D}_u$ are then propagated forward through $h_{\text{AP}}$. \\
        The CE loss in Eq. (1) is replaced by $l_{\text{DML}}$ in Eq. (5). \\
         $\mathcal{L}_{cls}= \mathcal{L}_x(\{{\tilde{\bm{y}_i}},\bm{p}_i\})+\mathcal{L}_x(\{{\tilde{\bm{y}_i}},\bm{p}_i^{\prime}\})$ \\
         $ \ \ \ \ \ \ \ \ \ \ \ \  + \lambda_u \mathcal{L}_u(\{{\text{Sharpen}(\bm{p}_i,T)},\bm{p}_i^{\prime}\})$ \\
        \tcp{estimation of distribution}
        $\bm{\pi} = m \bm{\pi}+(1-m) \frac{1}{|\mathcal{B}|} \sum\nolimits_{\bm{x}_i \in \mathcal{B}} \bm{p}_i $  \\
        calculate $\mathcal{L}_\text{mix},\mathcal{L}_\text{cr}$ \\
        minimize loss $\mathcal{L}_{total}=\mathcal{L}_{cls}+\gamma(\mathcal{L}_{cr}+\mathcal{L}_{mix})$ 
    }
\end{algorithm}

\bibliographystyle{named}
\bibliography{ijcai23}

\end{document}